\documentclass[runningheads]{llncs}
\usepackage{graphicx}
\usepackage{caption}

\begin{document}
\title{Improving Deep Image Clustering With Spatial Transformer Layers}
\author{Thiago V.M. Souza\inst{1}\orcidID{0000-0002-9934-2283} \and \\
Cleber Zanchettin\inst{1}\orcidID{0000-0001-6421-9747}*}
\authorrunning{Thiago V.M. Souza \and Cleber Zanchettin}
\titlerunning{Improving Deep Image Clustering With ST Layers}

\institute{\inst{1}Universidade Federal de Pernambuco\\ Centro de Inform\'atica\\ Recife, Brazil\\ \email{tvms@cin.ufpe.br, cz@cin.ufpe.br}}
\maketitle              % typeset the header of the contribution
\begin{abstract}
Image clustering is an important but challenging task in machine learning. As in most image processing areas, the latest improvements came from models based on the deep learning approach. However, classical deep learning methods have problems to deal with spatial image transformations like scale and rotation. In this paper, we propose the use of visual attention techniques to reduce this problem in image clustering methods. We evaluate the combination of a deep image clustering model called Deep Adaptive Clustering (DAC) with the Spatial Transformer Networks (STN). The proposed model is evaluated in the datasets MNIST and FashionMNIST and outperformed the baseline model.

\keywords{Image Clustering \and Deep Neural Networks \and Visual Attention \and Spatial Transformer Networks \and Adaptive Clustering}
\end{abstract}
\section{Introduction}
The clustering task consists of dividing a set of data into subgroups where elements belonging to the same group are similar to each other and different from the elements of the other groups. Clustering is a method of unsupervised learning and is a common technique for statistical data analysis. 

In some cases, clustering is even important to supervised learning. In many real applications of large-scale image classification, the labeled data is not available or is not enough to train supervised models, since the tedious manual labeling process requires a lot of time and labor. A widely used strategy is to applying clustering to the unlabeled training data to group in similar instances and then use minimal human effort to label annotation based in the group elements. 

Image clustering is a challenging task due to the image intra-class variability. For a long time, classic techniques  such as K-means were the best option to image clustering \cite{application:7conf_kmeansfast}\cite{application:8jor_kmeansopt}. 
In recent years, deep neural networks have proved to be very effective in several image processing areas and deep clustering approaches reached the state-of-the-art in manifold image benchmarks using methods such as Deep Clustering Network (DCN) \cite{DCN}, Joint Unsupervised Learning (JULE) \cite{jule}, Deep Embbeded Cluster (DEC) \cite{dec}, Deep Embbeded Cluster with Data Augmentation (DEC-DA) \cite{decda} and Deep Adaptive Clustering (DAC) \cite{deepadaptiveclustering}.

The deep neural networks are extremely powerful. However, it has some problems with spatial image transformations like scale and rotation. The majority of Convolutional Neural Networks (CNN) typically employ max-pooling layers using small pooling regions (e.g., 2 x 2 or 3 x 3 pixels) to deal with image transformations. The max-pooling approach provides a spatial invariance of up to only a small region, and the intermediate feature maps in the CNN is not invariant to intra-class differences of the input data. 

Advanced techniques have been proposed to deal with this problem, such the visual attention solutions as Spatial Transformer Networks (STN) \cite{spatialtransformernetworks}. This modules can be inserted into the CNN as a layer and provides the ability to learn invariance to scale, rotation and the more general image deformations. We refer to these mechanisms as Spatial Transformer layers (ST layer).

In this paper, we investigate the use of a visual attention technique in deep clustering models to making the network more invariant to intra-class differences of the input data. To evaluate this approach, we added ST layers into the Deep Adaptive Clustering (DAC) \cite{deepadaptiveclustering} model. We have not found in the literature deep image clustering models that use ST layers in their composition to deal with intra-class variance. 

We evaluate our approach performing experiments with the MNIST \cite{mnist} and FashionMNIST \cite{fashionmnist} datasets. The next section reviews the related work, specifically on deep image clustering. In Section 3, we detail the background of the combined methods. Section 4 details the proposed approach Spatial Transformer - Deep Adaptive Clustering (ST-DAC). In Section 5, we present the experiments and Section 6 presents the final remarks. 

\section{Related Works}

Many works in deep image clustering have achieved remarkable results or have become important approaches in how to handle the clustering problem \cite{surveydeepclustering}. All of these methods are directly related to our proposal. Among these works, we highlight the Deep Clustering Network (DCN) \cite{DCN}, which combines a pretrained Autoencoder (AE) network with the k-means algorithm. The Joint Unsupervised Learning (JULE) \cite{jule} which uses a hierarchical clustering module and a CNN to generate the representations of the images;  each previous method joint optimize deep representations generation and the function to build image clusters. Other interesting models are based on Generative Adversarial Networks (GAN)\cite{gan} and Variational Autoencoder (VAE) \cite{vae} as Categorical Generative Adversarial Networks \cite{catgan} and Variational Deep Embedding \cite{vade}. The models generate new images related to the learned image groups, besides performing clustering of these images.

The Deep Adaptive Clustering (DAC) \cite{deepadaptiveclustering} is historically one of the most representative methods in this category. Another method that brings much attention to the deep image clustering literature is the Deep Embbeded Clustering (DEC) \cite{dec}. The method performs a pretraining on a Stacked Autoencoder, then arranges the layers of the architecture to form a Deep-Autoencoder, in which the fine-tuning is performed. Then the part of the decoder is removed of the network, and the output of the encoder serves as the feature extractor for the clustering module. The network is optimized using the hardness-loss clustering method to assign the labels to the samples iteratively. This model is a reference for the evaluation of new models and experiments with deep image clustering.

All these models can achieve interesting results proposing modifications in the clustering functions, autoencoder, as well as in the network optimization. However, these works do not focus on problems already known in the deep learning approaches, such the difficult to deal with input spatial transformations variance.

The model Deep Embedded Cluster with Data Augmentation (DEC-DA) \cite{decda} follow a different approach and seeks to improve the generalization capacity of the network. In this model, the authors first train an AE in which the inputs are images with data augmentation. With the network trained and able to generate representative features, the clustering cost function and the reconstruction function of the features of the combined AE are employed in training. In this process, the decoder generates the data augmentation images features. The centroids are calculated from the representations generated by the same decoder; however, in this step, the decoder receives the original images without transformations. The whole network is optimized together. The use of data augmentation to build the AE proved to be quite efficient and reached state-of-the-art results in several datasets. Data augmentation is a technique capable of improving the generalization of the networks, however, the models still encounter difficulties when dealing with images transformations beyond those found in the augmented samples.

\section{Preliminaries}

\subsection{Deep Adaptive Clustering - DAC}\label{AA}

DAC \cite{deepadaptiveclustering} is a model of deep image clustering, based on a single-stage CNN, i.e. to perform the clustering of the images it is not necessary pretraining stages, nor additional stages of sequential independent clustering modules.

The model presents a somewhat innovative approach when dealing with the clustering problem, dealing with the problem differently of the other deep image clustering models. Usually, the methods use more specific clustering techniques such as hierarchical clustering or K-means. 
DAC proposes to attend the clustering task as a pair binary classification. In this way, the pairs of images are considered as belonging to the same cluster, or different clusters, depending on their similarities. 

\begin{figure*}[htbp]
\centerline{\includegraphics[scale=0.2]{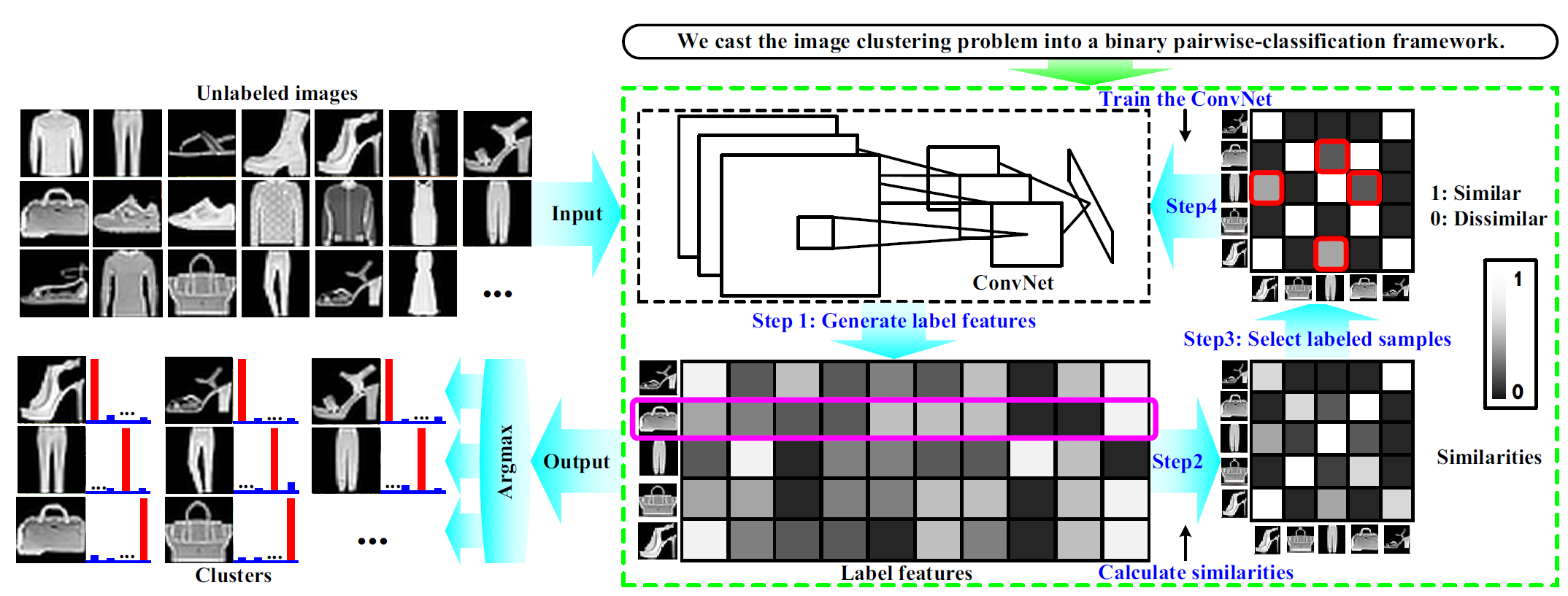}}
\caption{Diagram of the Deep Adaptive Clustering solution. Image adapted from \cite{deepadaptiveclustering}}
\label{DAC}
\label{fig}
\end{figure*}

In a classification approach, we need the labels of the classes to train the model. In order to overcome this situation in a clustering approach, the method obtains the labels from the CNN extracted features. The technique employs some constraint on the output of the model and thus manages to generate features suitable for clustering, making the learned feature labels tending to one-hot vectors. Moreover, it uses an adaptive learning algorithm for tuning the model. Figure~\ref{DAC} presents all steps of the adaptive training method.

The CNN in the early stages of the training process is still not able to distinguish very well the similarity between images with high intra-class variance. In this way, it is necessary to employ an adaptive method for selecting the training examples, choosing samples with a high level of confidence in the predictions. This method consists of firstly select training labels from the pairs of images that show similarities or dissimilarities verified by the network during the first training iterations. These similarities correspond with a higher probability that the result of the classification between the pair is correct. As the training progresses, the network becomes more robust, so samples of training pairs with a higher level of uncertainty are presented gradually. Thus the network learns more refined clustering patterns. Based on the similarity between the pairs, the training labels are generated as show Equation ~\ref{eqn:labeldac}.

\begin{equation}
\label{eqn:labeldac}
r _ { i j } = \left\{ \begin{array} { l } 

{ 1 , \;\; \ if  \;\; f \left( \mathbf { x } _ { i } ; \mathbf { w } \right) \cdot f \left( \mathbf { x } _ { j } ; \mathbf { w } \right) \geq u ( \lambda ) } \\

{ 0 , \;\;\ if  \;\;f \left( \mathbf { x } _ { i } ; \mathbf { w } \right) \cdot f \left( \mathbf { x } _ { j } ; \mathbf { w } \right) < l ( \lambda ) , \quad i , j = 1 , \cdots , n , } \\

{ Nothing, \; in \; other \; case } \end{array} \right.
\end{equation}

\noindent
Given $\mathcal { X } = \left\{ \mathbf { x } _ { i } \right\} _ { i = 1 } ^ { n }$ as the set of non-labeled images presented to the cluster where the variable ${x_{i}}$ is the i-th image of the data set. $x_{i}$ and $x_{j}$ are different unlabeled input images and $r_{ij}$ is an unknown binary output variable that receive the training label generated, where if $r_{ij} = 1$ the images $x_{i}$ and $x_{j}$ belong to the same group, in other case, if $r_{ij} = 0$, the images belong to different groups. $w$ are the actual parameters from the network; $f$ is a mapping function that maps input images to label features and $f \left( \mathbf { x } _ { i } ; \mathbf { w } \right) \cdot f \left( \mathbf { x } _ { j } ; \mathbf { w } \right)$ represents the dot product between two label features. $\lambda$ is a adaptive parameter that controls the selection of samples presented for training, $u(\lambda)$ is the threshold for selecting samples of similar pairs, $l(\lambda)$ is a threshold used to select samples of dissimilar pairs and $ `` $ Nothing $ "$ means that the sample $ (xi, xj, r_{ij}) $ is not presented for the training process.

DAC stops with the use of all training instances, and the objective cannot be improved. The DAC optimization function is defined as in the Equation~\ref{eqdac1}.

\begin{equation}
\min _ { \mathbf { w } } \mathbf { E } ( \mathbf { w } ) = \sum _ { i , j } v _ { i j } L \left( r _ { i j } , f \left( \mathbf { x } _ { i } ; \mathbf { w } \right) \cdot f \left( \mathbf { x } _ { j } ; \mathbf { w } \right) \right)
\label{eqdac1}
\end{equation}

In Equation~\ref{eqdac1} $v$ is an indicator coefficient where $v_{ij} = 1$ indicates that the sample is selected for training, and $v_{ij} = 0$ otherwise, $L$ is the function loss defined in Equation~\ref{eqdac2} where $ g \left( \mathbf { x } _ { i } , \mathbf { x } _ { j } ; \mathbf { w } \right) = f \left( \mathbf { x } _ { i } ; \mathbf { w } \right) \cdot f \left( \mathbf { x } _ { j } ; \mathbf { w } \right)$.

\begin{equation}
\begin{array} { l } { L \left( r _ { i j } , g \left( \mathbf { x } _ { i } , \mathbf { x } _ { j } ; \mathbf { w } \right) \right) = } \\ { - r _ { i j } \log \left( g \left( \mathbf { x } _ { i } , \mathbf { x } _ { j } ; \mathbf { w } \right) \right) \- \left( 1 - r _ { i j } \right) \log \left( 1 - g \left( \mathbf { x } _ { i } , \mathbf { x } _ { j } ; \mathbf { w } \right) \right) } \end{array}
\label{eqdac2}
\end{equation}

Finally, the model cluster the images according to the most significant label features. The DAC reached the state-of-the-art in several public datasets \cite{deepadaptiveclustering}.

\subsection{Spatial Transformer Networks - STN}\label{BB}

The Spatial Transformer Networks (STN)\cite{spatialtransformernetworks} is a visual attention mechanism consisting of differentiable modules, which can be trained using the backpropagation algorithm. The model can learn to perform spatial transformations conditioned to the input data mapping.

\begin{figure}[!htbp]
\centerline{\includegraphics[scale=0.25]{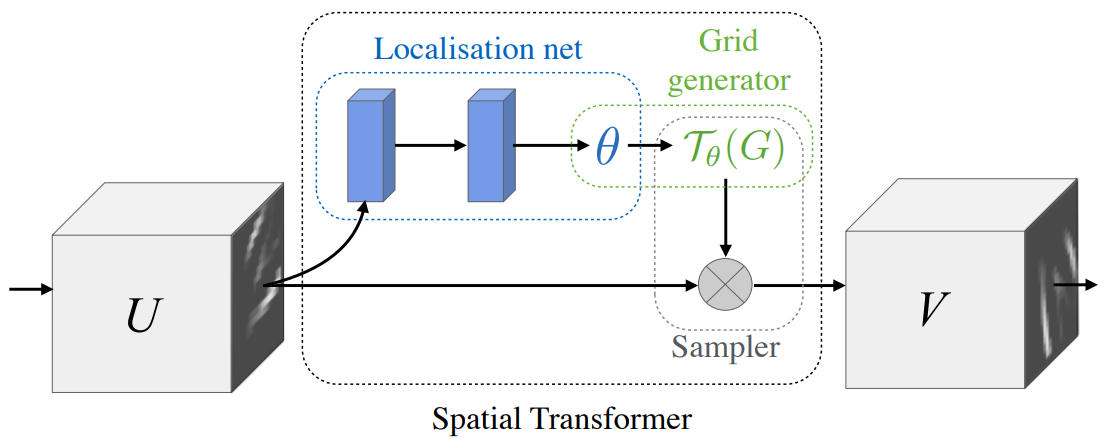}}
\caption{Spatial Transformer Network. Source: \cite{spatialtransformernetworks}}
\label{figstn}
\end{figure}

The STN modules can be inserted as layers of a CNN and are composed of three distinct parts, as we can see in Figure ~\ref{figstn}, the localization network, the grid generator, and the sampler. The network receive the input data map $U \in \Re ^ { H \times W \times C }$ with $W$, $H$, $C$ are the width, height, and channels, respectively. The input information can be the input image or the features map extracted by the inner layers of a CNN. 

The $\theta$ transformation parameters are predicted by the localization network, which can be any CNN or Multi-Layer Perceptron with a regression layer at the output. The size of $\theta$ may vary according to the desired transformation type.

The transformation parameters are then used to generate a transformation matrix $\tau _ { \theta }$. This matrix is applied in a sampling grid $G$ produced by the grid generator. The sampling grid is composed of normalized coordinates that map the access to each input feature map values. Finally, the sample kernel uses the grid and the mapped features to generates the output map $U$.

Through these mechanisms, convolutional networks can become more robust and invariant to the transformations or variability inherent in the input images and with a low computational cost.

\section{Proposed pipeline}\label{CC}

The method proposed in this work aims to create a new approach by adding to the DAC model some ST layers for deep image clustering. We call this new model Spatial Transformer - Deep Adaptive Clustering (ST-DAC).

To evaluate our hypothesis, we use the DAC* model, a more simple DAC version also presented in the original DAC paper \cite{deepadaptiveclustering}. In the DAC* the upper and lower sampler selection thresholds are set by the parameter $\lambda$ that is added linearly at each epoch. In this method, at each iteration, all examples are also selected for training.

The CNN that composes the architecture present in the original paper is an AllConvNet \cite{allconv}. However, in several experiments, we had difficulties in training the model using the ST layers. In these cases, the ST layers performed strange transformations in the images, distancing the object and making the input image noisily after some epochs and impairing the results. This behavior led us to believe that the problem could be due to the vanishing gradient. Another option could be to find an appropriate learning rate that attended the training of the CNN and STN layers at the same time. 

\begin{figure*}[!htbp]
\centerline{\includegraphics[scale=0.3]{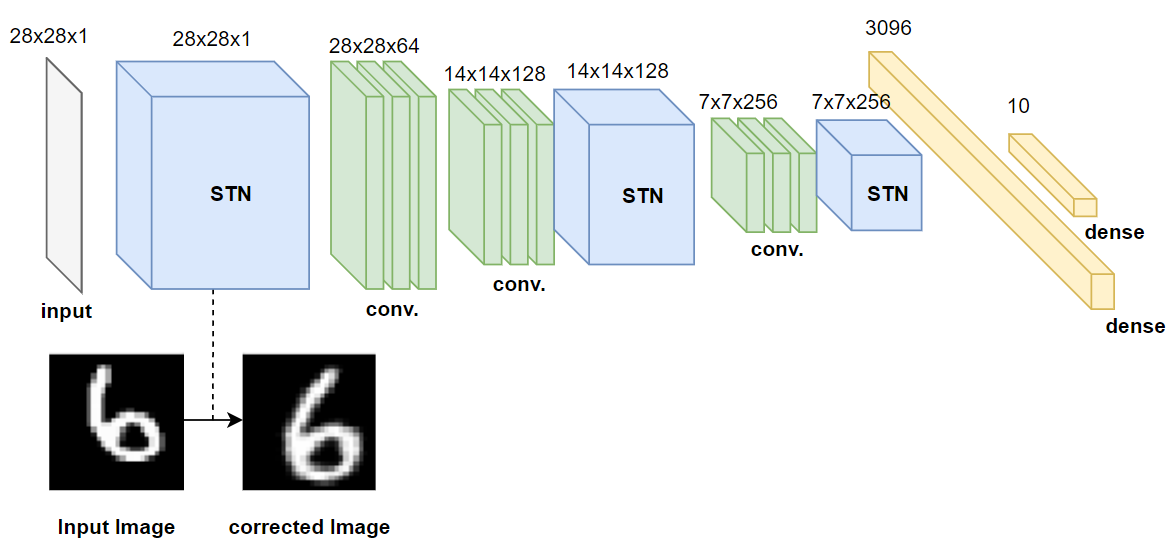}}
\caption{The proposed convolutional architecture has three spatial transformer layers. The first one is inserted after the input layer and performs transformations in the initial image. The other ST layers are applied in the feature maps after the second and third block of convolutional layers.}
\label{figProposed}
\end{figure*}

To minimize this problem, we replace the standard CNN for a smaller model based on the VGG network \cite{vgg}. The new model has similar results to the original approach presented in DAC. In Table~\ref{table1} are presented the parameters of the proposed CNN. We verify different arrangements of ST layers, we inserted the layer after the input image and also after the convolutional layers. Finally, we used 3 ST layers, and Figure~\ref{figProposed} shows the complete network architecture.

We inserted one ST layer after each block of convolutional layers, applying corrections to the original input data, like in \cite{spatialtransformernetworks}, and we perform transformations in the extracted features when it suffers a significant spatial transformation as after max-pooling downsampler. This structure immediately corrects the image transformation experienced by the input data, as we can see in the Figure~\ref{figProposed}. The model does not use an ST layer after the first convolutional layers block, to reduce the computational cost, because this block does not perform resolution changes. The structure of the localization network in the spatial layers is the same proposed in \cite{DSTN} and detailed in Table~\ref{table2}. Usually, the localization network structures use the Relu activation function between the layers and leave the last layer without activation \cite{spatialtransformernetworks}. Unconventionally, \cite{DSTN} uses tanh activation function after all the convolutional and dense layers. In the initial experiments, it showed better results with the ST layers than the conventional structures. 

\begin{multicols}{2}

\captionof{table}{The architecture of the VGG based used in ST-DAC.}
\begin{center}
%\vspace{2mm}
\begin{tabular}{c}
\hline
\hline
\textbf{Model VGG Based}\\
\hline
\hline
Input 28x28 monochrome image \\
\hline
3 x 3 conv. 64 BN ReLU \\
\hline
2 x 2 Maxpooling BN \\
\hline
3 x 3 conv. 128 BN ReLU \\
\hline
2 x 2 Maxpooling BN \\
\hline
3 x 3 conv. 256 BN ReLU \\
\hline
2 x 2 Maxpooling BN \\
\hline
3096 dense BN ReLU \\
\hline
10 dense BN ReLU SoftMax \\
\hline
                              \\
\end{tabular}
\label{table1}
\end{center}

\columnbreak

\caption{The architecture of the Localization Network utilized in ST Layers.}

\begin{center}
%\scriptsize
\begin{tabular}{c}
\hline
\hline
\textbf{Localization Network}\\
\hline
*Input NxNxM monochrome image \\
\hline
2 x 2 Maxpooling \\
\hline
5 x 5 conv. 20 Tanh \\
\hline
2 x 2 Maxpooling \\
\hline
5 x 5 conv. 20 Tanh \\
\hline
50 dense Tanh \\
\hline
6 dense Tanh \\
\hline
$^{\mathrm{*}}$At the input NxNxM corresponds\\ the output dimensions of the anterior\\ layers to ST module.
\end{tabular}
\label{table2}
\end{center}

\end{multicols}

\section{Experiments}\label{DD}
The experiments were conducted with two well-used datasets to evaluate computer vision problems: the MNIST\cite{mnist} handwritten digits and the Fashion MNIST\cite{fashionmnist} based on clothing images. 

The metrics used to evaluate the methods are the Adjusted Rand Index (ARI), Normalized Mutual Information (NMI), and Clustering Accuracy (ACC). These measures return results between a range [0,1], values close to 1 represent more precise results in clustering.

\subsection{Experimental Settings}\label{GG}

To evaluate the performance of the proposed ST-DAC model we perform experiments comparing the proposed approach with DEC \cite{dec}, DCN \cite{DCN}, VADE \cite{vade}, JULE \cite{jule}, DEC-DA \cite{decda}. The most outstanding approach in the literature is ConvDEC-DA \cite{decda}, we also compare our results with this method. Besides, we also compare the proposed model with the DAC \cite{deepadaptiveclustering} and its original version DAC*. These models present the best literature results in the two evaluated datasets.

In order to verify the importance of the ST layers to the proposed model performance, we evaluate our approach with four experiments: 1) with all the ST layers activated (ST-DAC + 3 ST Layers); 2) with the last layer off (ST-DAC + 2 ST Layers); 3) with the first layer activated (ST-DAC + 1 ST Layers); and 4) with no ST layers (ST-DAC without ST Layers). The idea is to quantify the contribution of the ST layers in the model's accuracy. 

The original DAC \cite{deepadaptiveclustering} paper does not perform experiments in the Fashion MNIST dataset. For comparison purposes, we run the DAC* version in the Fashion MNIST and MNIST datasets using the same parameters suggested in the original paper. We also used the same data augmentation parameters proposed in the original DAC paper in both datasets. We only modified the initial lower and upper selection thresholds to the range [0.9 0.99] for the MNIST dataset and [0.8 0.99] for the Fashion MNIST dataset. We used Adam optimizer in our model with a learning rate of 0.0001, as suggested in \cite{DSTN}. We run each experiment 10 times and calculate the average of the results, which is used for comparison with the other methods.

Table~\ref{tabResultsreference} presents the experiments. The results taken from literature, are marked with the symbol $\dagger$, the references are next to the name of the method. Results tagged with - are not available. We evaluated the behavior of STNs during the whole training phase in order to verify its contribution to the effectiveness of the model. For this purpose, the results obtained by the ST-DAC during the training were sampled, and the images of the first STN layer were extracted from the model to verify the quality of the spatial transformations applied in the input image during the training steps. 

The source code of the experiments is available at a public repository \footnote{https://github.com/tvmsouza/ST-DAC.}. 

\section{Results}

The results of the experiments with the evaluated methods are presented in Table ~\ref{tabResults}. The proposed model ST-DAC without the ST layers presented inferior performance to those obtained by the DAC model. This is a familiar scenario because, compared to the original DAC model, the ST-DAC model uses a shallow convolutional network to extract the image representations. However, we can observe 
the ST-DAC model trained with and without ST layers was executed on the same configuration parameters. Thus, it suggests that the gain in performance and results of the model resulted from the use of the ST layers in its convolutional network.

Using only one ST layer after the input layer of the proposed model, we obtain a superior result in almost all metrics in the two datasets, compared to the best results previously obtained by DCN, JULE, VADE, DEC, DAC* and DAC.

Adding one ST layer before the input layer and one ST layer after the extracted features of the second block of convolutional layers, we were able to surpass the previous best results of DCN, JULE, VADE, DEC, DAC * and DAC in both datasets and surpassing the ConvDEC-DA and DEC in Fashion MNIST by a large margin of accuracy.

\begin{table}[!htbp]
\caption{Clustering performance of DAC* and the proposed ST-DAC method, obtained from our experiments and compared methods results taken from the literature marked with $\dagger$, considering Clustering Accuracy (ACC), Normalized Mutual Information (NMI) e Adjusted Rand Index (ARI).}
\begin{center}
\scriptsize
\begin{tabular}{|c|c|c|c|c|c|c|}
\hline
\textbf{}&\multicolumn{6}{|c|}{\textbf{Datasets}} \\
\cline{2-7} 
\textbf{Model}&\multicolumn{3}{|c|}{\textbf{MNIST}}&\multicolumn{3}{|c|}{\textbf{Fashion MNIST}} \\
\cline{2-7} 
\textbf{} & \textbf{\textit{ACC}}& \textbf{\textit{NMI}}& \textbf{\textit{ARI}}& \textbf{\textit{ACC}}& \textbf{\textit{NMI}}& \textbf{\textit{ARI}} \\
\hline
 DCN\cite{decda}$\dagger$&0.830&0.810&-&0.501&0.558&- \\
\hline
 VADE\cite{decda}$\dagger$&0.945&0.876&-&0.578&0.630&- \\
\hline
 JULE\cite{decda}$\dagger$&0.964&0.913&-&0.563&0.608&- \\
\hline
 DEC\cite{decda}$\dagger$&0.863&0.834&-&0.518&0.546&- \\
\hline
 ConvDEC-DA\cite{decda}$\dagger$&\textbf{0.985}&\textbf{0.962}& - &\textbf{0.586}&\textbf{0.636}& - \\
\hline
 DAC\cite{deepadaptiveclustering}$\dagger$&0.977&0.935&0.948&-&-&-   \\
\hline
 DAC*\cite{deepadaptiveclustering}$\dagger$&0.966&0.924&0.940&-&-&-   \\
\hline
\hline
DAC*&$0.974$&$0.944$&$0.945$8&$0.628$&$0.589$&$ 0.483$   \\
\hline
ST-DAC without ST Layers&$0.957$& $0.932$ &$0.923$ &$0.612$ &$0.591$ &$0.460$ \\
\hline
ST-DAC + 1 ST Layer&$0.978$ &$0.950$ &$0.953$ &$0.649$ &$0.650$ &$0.520$    \\
\hline
ST-DAC + 2 ST Layers&$\textbf{0.980}$ &$\textbf{0.953}$&$\textbf{0.956}$&$0.656$&$0.658$&$0.533$   \\
\hline
ST-DAC + 3 ST Layers&$0.961$ &$0.936$ &$0.927$ &$\textbf{0.664}$ &$\textbf{0.668}$&$\textbf{0.541}$ \\
\hline

\end{tabular}
\label{tabResults}
\end{center}
\label{tabResultsreference}
\end{table}
 
It is notable that with the addition of more ST layers, the model can get better results. However, the idea of using more layers to improve results cannot be applied in all contexts. Using three ST layers in the Fashion MNIST experiments, we get the best result in this data set. However, in the MNIST experiments, the use of a third ST layer in some way compromised the data, reducing its final result compared to the network with two ST layers. 

\begin{figure}[htp]
\centerline{\includegraphics[scale=0.6]{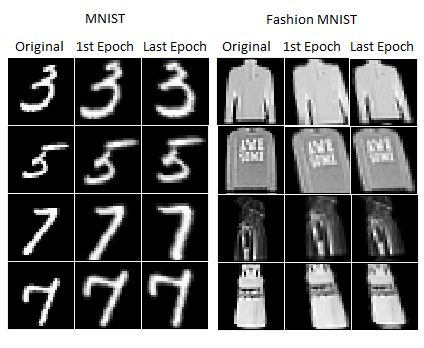}}
\caption{Comparison between some original images and their respective output from the first ST layer. In the left column is the original image, in the center the output of the first ST layer after the first training epoch and in the right column the output of this same layer after the training.}
\label{figcorrect}
\end{figure}

Figure~\ref{figcorrect} show some qualitative results, we can observe the extracted images from the output of the first transformation layer during different training epochs. We observe that the STNs presented the same behavior as the most STN literature works. The network initially applies a powerful zoom in the image and then reduces this zoom to fit the object in a region where it can better frame all the details of the area of interest, normalizing the objects, correcting distortions and rotation during the network training. In the training phase, we used data augmentation, and the ST layers learned how to correct various transformations of rotation, scale, and translation. With this knowledge, the ST layers rotate the objects to a standard angle in which it is possible to enlarge and fill a larger area of the entire image without losing relevant information of the target object. The output images of the ST layers also have a blurred aspect and lose some details, but this loss is offset by the transformation corrections defined above.

\begin{figure}[htp]
\centerline{\includegraphics[scale=0.6]{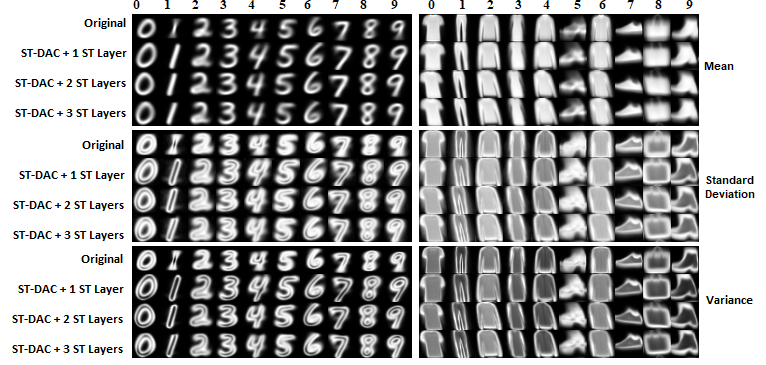}}
\caption{Average, standard deviations and variances from original and extracted images from the first ST layers inserted in ST-DAC models.}
\label{figStat}
\end{figure}

In Figure ~\ref{figStat} are exhibit the average, standard deviations, and variances between all images of the same class, from the original images and the processed images by the first ST layer of the ST-DAC models. All images were obtained after the training phase. It is possible to observe, in the averages results from MNIST, that the ST layers moved the image to a common region, which is evidenced by the high-density level values of the digits in the images. This normalization allows a better comparison between the elements. In addition, the STNs, as seen in previous images, applied a reliable approximation of the digits, which may have improved the capture of details and consequently, improved the results. Analyzing the standard deviation and the variance, it is possible to verify the thickness of the edges of the digits. We observed in the most cases that in the use of ST layers these thicknesses decreased, which indicates a more significant normalization between the positions and transformations of the objects from the same class. Besides, the internal gaps was a high density of variance, and the standard deviation decreased. 

We verify a similar behavior in experiments with the Fashion MNIST dataset. During training, the network learns from all these transformations variations, increasing its generalization as if the training is performing a data augmentation in a controlled way.

c

After a few first epochs, the results are better in ST-DAC with ST layers. It is possible to notice that, even before converging, the networks also benefit from the intermediate representations obtained by the ST layers during the training. The curves remained stable in the MNIST data set following the same growth pattern without significant declines in performance over time.

Analyzing the results of the Fashion MNIST dataset, we see that ST-DAC + 3 ST layers lose performance at some points between epochs two and six and then stabilize and continue to increase. Among the combinations of layers, the growth of the ST-DAC + 2 ST layers remained more stable in comparison to the others configurations during all the training periods, despite being overcome in the final result by the network with 3 ST layers. From all these results, it is noticeable that visual attention techniques allow models using simpler convolutional networks to obtain superior results to other methods of deep image clustering.

\section{Conclusion}

In this work, we propose a new approach to Deep Adaptive Clustering, replacing the original convolutional feature extraction of the DAC network with a new simpler model based on Spatial Transformer layers. We evaluated our approach by conducting experiments on two public datasets and compared it with other promissory literature methods to the problem.%\\[-8mm]

We also conducted experiments by varying the amount of ST layers in the proposed convolutional model, to evaluate if, with the addition of new ST layers, using the spatial transformation correction over the internal extracted features, the model performance grow proportionally. The experiments showed that our approach was able to outperform other methods in the two evaluated datasets, achieving state-of-art results in both datasets.

\begin{figure}[htp]
\centerline{\includegraphics[scale=0.4]{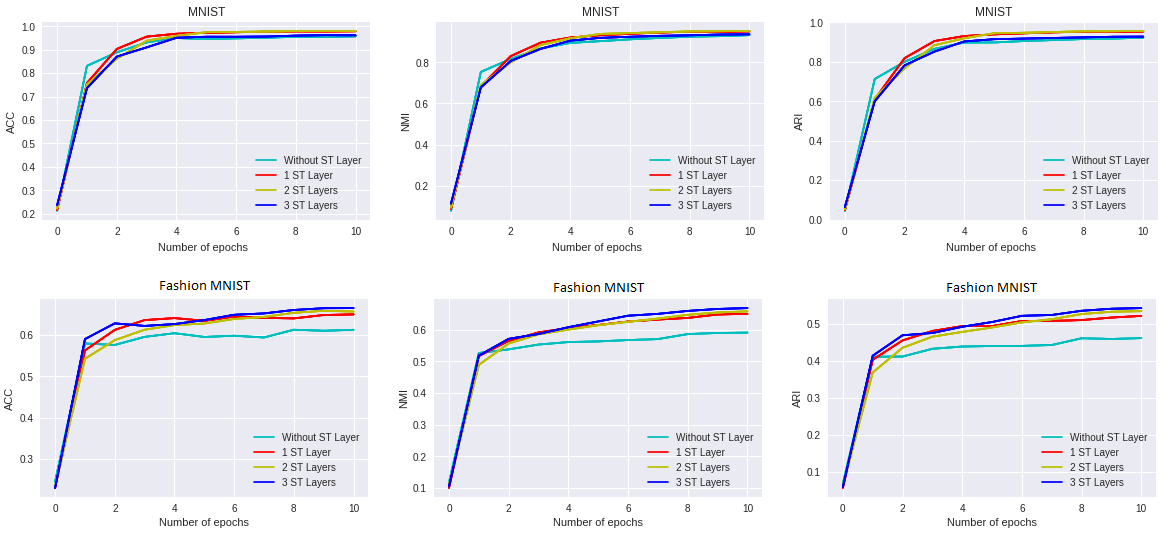}}
\caption{Comparison of clustering performance between models with diferent numbers of ST Layers during training epochs on MNIST(Top) and Fashion MNIST(Bottom).}
\label{fig7}
\end{figure}

Finally, we show that with the use of visual attention techniques, such as ST Layers, the deep image clustering method can obtain performance improvements. The use of ST layers has shown promising results to improve the performance of the DAC model. However, the area of visual attention continues to advance and several new approaches were proposed that extend the capacity of standard ST layers. We suggest for future work verify the performance of these new methods to improve deep image clustering models.

\section*{Acknowledgment}
This work was supported by CNPq (Brazilian research agency). We gratefully acknowledge the support of NVIDIA Corporation with the donation of the Titan XP GPU used for this research. %\\[-8mm]

%

\end{document}